\documentclass[11pt]{article}
\usepackage[utf8]{inputenc}
\usepackage[T1]{fontenc}
\usepackage{newtxtext,newtxmath}
\usepackage[letterpaper,margin=1in]{geometry}
\usepackage{amsmath}
\usepackage{booktabs}
\usepackage{array}
\usepackage{float}
\usepackage{fancyvrb}
\usepackage{xcolor}
\usepackage{tikz}
\usepackage{graphicx}
\usetikzlibrary{arrows.meta,positioning,fit,backgrounds}
\usepackage{pgfplots}
\pgfplotsset{compat=1.17}
\usepackage[labelfont=bf,labelsep=colon,font=small]{caption}
\usepackage{cite}
\usepackage[hidelinks]{hyperref}
\usepackage{url}
\begin{document}

\begin{center}
{\LARGE\bfseries Redacting Personal Data: A Speed and Accuracy Comparison of Regex, Presidio and Hybrid Designs\par}
\vspace{1.4em}

\noindent\parbox[t]{0.48\linewidth}{\centering\textbf{Sai Yashwant}\\ \small AI Product \& Platform Head\\ ManpowerGroup Services India Pvt. Ltd.\\ \footnotesize sai.yashwant@manpowergroup.com}\hfill\parbox[t]{0.48\linewidth}{\centering\textbf{Manoj Kumar Gondimalla}\\ \small AI Operations Lead\\ ManpowerGroup Services India Pvt. Ltd.\\ \footnotesize manojkumar.gondimalla@manpowergroup.com}\par\vspace{1em}
\noindent\parbox[t]{0.48\linewidth}{\centering\textbf{Arjun Sharma}\\ \small AIOps Engineer\\ ManpowerGroup Services India Pvt. Ltd.\\ \footnotesize arjun.sharma@manpowergroup.com}\hfill\parbox[t]{0.48\linewidth}{\centering\textbf{Shubham}\\ \small Data Scientist\\ ManpowerGroup Services India Pvt. Ltd.\\ \footnotesize shubham.kumar@manpowergroup.com}\par\vspace{1.2em}

\end{center}
\begin{center}\textbf{ABSTRACT}\end{center}
\begin{quote}
Many enterprise projects now pass free text to large language models (LLMs), and that text often contains personally identifiable information (PII) such as names, phone numbers, email addresses, home addresses and dates of birth. Once a prompt is sent, this data is copied into provider systems, application logs and observability traces. Redacting PII before the call is an obvious control, yet teams seldom have numbers on what it costs in latency or what it lets through. This paper compares four redaction designs that any LLM pipeline can adopt: a pure regular-expression (regex) redactor, Microsoft Presidio running on a spaCy model, a hybrid in which Presidio finds only person names while regex handles structured identifiers, and a fast hybrid that calls spaCy's named-entity component directly. All four share a known-entity rule, which removes the person the system already knows it is handling, and a consistent placeholder scheme, so the LLM can still tell one person from another after redaction. We evaluate the designs on a sample workflow, which we call ProfileAssist, that turns resumes into structured profiles, using 100 synthetic resumes (523 characters on average, with 900 PII values and 600 business terms that must survive). Regex ran at a 0.20 ms median but caught only 88.9\% of PII, because it cannot find names that are not known in advance, such as the references listed in a resume. Presidio, hybrid and fast hybrid each caught 100\% of PII and kept 100\% of business terms, at median latencies of 37.59 ms, 30.23 ms and 13.48 ms. Measured against a 9,120 ms median LLM call, even the slowest option adds only 0.41\% to the request. We therefore recommend the fast hybrid as a default for LLM pipelines that handle personal data, with regex kept as a safety net rather than as the only control.

\textbf{Keywords}\quad PII redaction, Microsoft Presidio, regular expressions, named entity recognition, spaCy, LLM privacy, LLMOps, data minimization
\end{quote}

\section{Introduction}
Large language models are now part of many enterprise projects. Teams use them to summarize documents, classify requests, extract fields from free text and draft replies. In a large share of these projects the input contains personally identifiable information (PII): a support ticket names the customer and gives a phone number, an HR document lists an employee's address, and a resume holds a person's contact details along with the names of their references. The value of the LLM is real, but so is the risk, because every prompt that carries PII spreads it further.

That data does not stay in one place once the call is made. It travels to the model provider, it is written into application logs, and in a well-run LLMOps setup it is also captured by an observability platform, where prompts and responses are stored as traces for debugging and evaluation. Each copy is one more place where personal data must be protected, retained and finally deleted under rules such as the GDPR~\cite{gdpr}, the CCPA~\cite{ccpa} and India's Digital Personal Data Protection Act~\cite{dpdp}. Research on training-data extraction and PII leakage shows that what goes into a language model, or into a dataset later used for tuning, can come back out~\cite{carlini,lukas}. For this reason the OWASP Top 10 for LLM applications lists sensitive information disclosure among the leading risks of LLM systems~\cite{owasp}.

The simplest defense is to redact PII before it ever reaches the model. Here, however, engineering teams face a familiar trade-off. Regular expressions are fast and predictable, and they handle structured identifiers such as emails and phone numbers well, but they cannot recognize a name they have never seen. Machine-learning tools such as Microsoft Presidio~\cite{presidio}, which relies on spaCy named entity recognition (NER)~\cite{spacy}, can find unseen names, yet they cost more CPU time and take seconds to load. In a request path that already waits on an LLM, it is not obvious whether that extra cost matters, and so teams often decide on instinct rather than on measurement.

This paper replaces that instinct with numbers. We compare four redaction designs that can be placed in front of any LLM call: regex alone, Presidio, a hybrid of Presidio names plus regex, and a fast hybrid that uses only spaCy's NER component plus regex. All four reuse one redaction core with a known-entity rule and consistent placeholders. We report accuracy both as PII recall and as business-term retention, so that over-redaction is measured as well as leakage, and we set load time and latency against the median latency of an LLM call. We then turn the results into practical guidance for any team that runs LLM pipelines over personal data.

To keep the measurements concrete, we ground them in a sample workflow that we call ProfileAssist. In ProfileAssist, a resume is passed to an LLM that returns a structured profile with skills, job titles and a short summary. We chose resumes because they are among the densest sources of PII a company handles, and because they show a pattern that recurs in many other projects. The text is about one person the system already knows, yet it also mentions other people, such as references, whom the system does not know. The same pattern appears in support tickets that mention agents, in case notes that mention relatives, and in email threads that copy colleagues. The example, in other words, uses resumes, but the problem is general.

ProfileAssist and all names in this paper are illustrative. All benchmark text is synthetic, and the only measured external figure is the median LLM latency, taken from traces of a representative LLM workload.

\section{Related Work}
Automatic removal of personal identifiers from free text has a long history in clinical informatics. Sweeney's Scrub system combined pattern templates with lists of known names to clean medical records~\cite{sweeney}, and later rule-based tools, such as the de-identification software of Neamatullah et al., relied on regular expressions, dictionaries and context rules~\cite{neamatullah}. These early systems already showed that patterns work well for formatted identifiers but struggle with names, which vary too much to list. Shared tasks such as the 2014 i2b2/UTHealth challenge then pushed the field toward machine learning, and hybrid systems that mixed rules with statistical models were often near the top of the rankings~\cite{stubbs}, and recurrent networks later reached high recall on patient notes without hand-crafted features~\cite{dernoncourt}.

Outside healthcare, most of the progress came from general-purpose NER. The BiLSTM-CRF model of Lample et al.~\cite{lample} and the transformer models that followed made person detection reliable enough to package in open libraries. spaCy offers fast, production-ready pipelines in which NER is one component next to a tagger, a parser and a lemmatizer~\cite{spacy}. Presidio builds on such pipelines and adds pattern recognizers, checksum validation and context words, together with an anonymizer that replaces the detected spans~\cite{presidio}. Lison et al. reviewed text anonymization more broadly and argued that detection quality must always be weighed against how much useful content survives~\cite{lison}. The Text Anonymization Benchmark (TAB) of Pilán et al. made the same point with an evaluation framework that scores privacy protection and utility side by side~\cite{tab}, and Yermilov et al. showed that pseudonymization, rather than blanket masking, helps downstream language tasks keep working~\cite{yermilov}.

Our work is narrower than these studies, but more operational. We do not propose a new detector. Instead, we take tools that teams actually deploy, run them on text shaped like a real LLM prompt, and measure what matters in a live request path: load time, tail latency, leakage and over-redaction. Few published studies set the cost of redaction against the cost of the LLM call it protects. Guidance such as NIST SP 800-122~\cite{nist122} and the NIST AI Risk Management Framework~\cite{nistrmf} asks organizations to keep PII exposure to a minimum, and this paper offers one measured way of doing so that is not tied to a single domain.

\section{Problem Setting}
\subsection{Redaction in a Generic LLM Pipeline}
In a typical LLM pipeline, a service gathers some text, wraps it in a prompt and sends the prompt to an LLM. Often the service already knows whose data it is handling, for example the customer behind a ticket or the employee behind a record, because that identity comes from its own systems. We call this person the known entity. The redactor sits between prompt assembly and the LLM client, as Figure 1 shows. Because redaction happens before the call, the model provider, the application logs and the trace store all receive the same placeholder text. The LLM sees tokens such as [PERSON\_1] or [PHONE\_NUMBER\_2] in place of real values, and the service can map them back afterwards if its output needs them.

\begin{figure}[H]
\centering
\resizebox{\linewidth}{!}{%
\begin{tikzpicture}[>=Stealth, font=\footnotesize,
  base/.style={draw, rounded corners=3pt, align=center, inner sep=4pt, line width=0.6pt},
  data/.style={base, fill=gray!8, text width=2.9cm},
  det/.style={base, fill=blue!10, text width=3.3cm, minimum height=1.0cm},
  merge/.style={base, fill=blue!25, text width=2.0cm, minimum height=3.4cm},
  ext/.style={base, fill=orange!12, text width=2.3cm, minimum height=1.3cm},
  arr/.style={->, line width=0.7pt}]

\node[data] (in) at (1.45,0) {\textbf{Raw prompt}\\ \scriptsize (example: resume)\\[3pt]
  \texttt{\scriptsize KNOWN\_SUBJECT:}\\
  \texttt{\scriptsize Andre Kowalczyk}\\
  \texttt{\scriptsize (901) 555-4821}\\
  \texttt{\scriptsize Ref: Greta Fontaine}};

\node[det] (d1) at (5.0,1.3)  {\textbf{1. Known-entity rule}\\ the person the system already knows};
\node[det] (d2) at (5.0,0)    {\textbf{2. Regex patterns}\\ email, phone, address, ZIP, DOB, URL, SSN};
\node[det] (d3) at (5.0,-1.3) {\textbf{3. spaCy NER}\\ names not known in advance (e.g.\ references)};

\node[merge] (m) at (8.4,0) {\textbf{Merge spans}\\[3pt] resolve overlaps,\\ assign consistent\\ placeholders};

\begin{scope}[on background layer]
\node[draw=blue!50, rounded corners=6pt, fill=blue!3, line width=0.6pt,
      fit=(d1)(d2)(d3)(m), inner sep=6pt,
      label={[font=\footnotesize\bfseries, anchor=south]north:PII redactor}] (red) {};
\end{scope}

\node[data, fill=green!8] (out) at (11.9,0) {\textbf{Placeholder prompt}\\[3pt]
  \texttt{\scriptsize KNOWN\_SUBJECT:}\\
  \texttt{\scriptsize [PERSON\_1]}\\
  \texttt{\scriptsize [PHONE\_NUMBER\_1]}\\
  \texttt{\scriptsize Ref: [PERSON\_3]}};

\node[ext] (llm)  at (15.3,1.1)  {\textbf{LLM task}\\ (e.g.\ profile summary)};
\node[ext] (logs) at (15.3,-1.1) {\textbf{Logs \& traces}\\ provider logs, app logs, observability};

\draw[dashed, red!60!black, line width=0.8pt] (10.0,2.6) -- (10.0,-2.6);
\node[anchor=north east, font=\scriptsize\itshape, text=red!60!black] at (9.9,-2.65) {real PII stays inside};
\node[anchor=north west, font=\scriptsize\itshape, text=red!60!black] at (10.1,-2.65) {only placeholders leave};

\draw[arr] (in.east) -- (d1.west);
\draw[arr] (in.east) -- (d2.west);
\draw[arr] (in.east) -- (d3.west);
\draw[arr] (d1.east) -- (m.west |- d1.east);
\draw[arr] (d2.east) -- (m.west |- d2.east);
\draw[arr] (d3.east) -- (m.west |- d3.east);
\draw[arr] (m.east) -- (out.west);
\draw[arr] (out.east) -- (llm.west);
\draw[arr] (out.east) -- (logs.west);
\draw[arr, dashed] (llm.south) -- (logs.north) node[midway, right, font=\scriptsize] {traced};
\end{tikzpicture}%
}
\caption{How redaction works in a generic LLM pipeline, shown with values from the ProfileAssist example. Three detectors find PII spans: the known-entity rule, regex patterns, and spaCy NER for names that are not known in advance (the regex-only design skips this step). The spans are merged and replaced with consistent placeholders, so only placeholder text crosses the service boundary to the LLM, the logs and the traces.}
\end{figure}
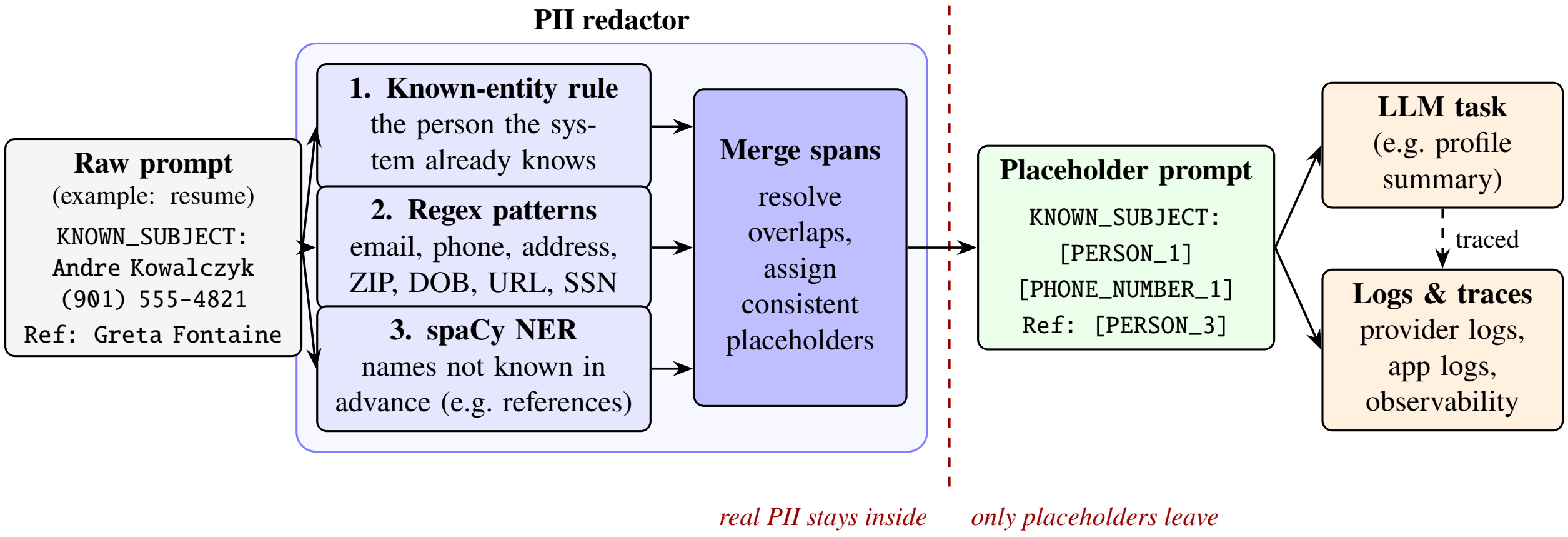

\subsection{Running Example: ProfileAssist}
ProfileAssist is a sample service that takes a person's resume and asks an LLM for a structured profile with skills, job titles, a short summary and a suggested job family. The known entity is the resume owner, and the prompt names them on a line of the form \texttt{KNOWN\_SUBJECT: \textless{}name\textgreater{}} before the resume, which sits inside \texttt{\textless{}subject\_data\textgreater{}} tags. As the latency yardstick we use a median (P50) LLM call time of 9,120 ms, observed in traces of a representative LLM workload. Whatever redaction adds should be small next to this figure.

\subsection{What Must Be Removed and What Must Stay}
Two kinds of error matter in any redaction setting. A leak happens when a PII value reaches the model untouched. Over-redaction happens when business content is masked by mistake, for example when a skill such as ``pallet jack'' is tagged as a person. A leak is a privacy failure, while over-redaction is a quality failure, because the LLM cannot use content it cannot see. Which terms count as business content depends on the project, and Table 1 lists the PII types and business terms for our example. City and state stay in the text on purpose: they are useful for the profile and, on their own, they do not identify a person. The last column shows which items are counted in the accuracy scores of Section 5.

\begin{table}[H]
\centering
\caption{PII types removed and business terms kept in the ProfileAssist example, with values from the synthetic corpus.}
\small
\begin{tabular}{>{\raggedright\arraybackslash}p{0.106\linewidth}>{\raggedright\arraybackslash}p{0.290\linewidth}>{\raggedright\arraybackslash}p{0.378\linewidth}>{\raggedright\arraybackslash}p{0.106\linewidth}}
\toprule
\textbf{Category} & \textbf{Item} & \textbf{Example} & \textbf{Scored} \\ \midrule
PII & Subject name (full name and parts) & Andre Kowalczyk, ANDRE KOWALCZYK, Andre & Yes \\
PII & Email address and phone number & andre.kowalczyk0@gmail.com, (901) 555-4821 & Yes \\
PII & Street address and ZIP code & 4521 Maple Ave, 38103 & Yes \\
PII & Date of birth & 04/17/1991 & Yes \\
PII & Reference name and phone number & Greta Fontaine, 412-555-7310 & Yes \\
PII & Profile URL and US SSN & linkedin.com/in/andrekowalczyk, 123-45-6789 & No \\
Keep & City, state and employer & Memphis, TN, Northwind Logistics & Yes \\
Keep & Skills and certifications & forklift operation, OSHA 10, pallet jack & Yes \\
\bottomrule
\end{tabular}
\end{table}

\section{Redaction Approaches}
All four designs share one core and differ only in how they find spans, so any difference in the results comes from detection and not from formatting.

\subsection{Shared Redaction Core}
The core has three parts. The first is a set of compiled regular expressions, listed in Table 2~\cite{friedl}. Some patterns use a named group called \texttt{pii} so that only part of a match is replaced. The ZIP pattern, for instance, matches ``TN 38103'' to make sure the number follows a state code, yet it replaces only ``38103'' and leaves the state readable. The date-of-birth pattern works in the same way and replaces a date only when it follows ``DOB'' or ``Date of birth'', so that employment years such as 2021 are left alone. Because our example data is US-based, the patterns target US formats; a project in another region would swap in its own, for instance Indian phone numbers or PAN and Aadhaar numbers.

\begin{table}[H]
\centering
\caption{Regular-expression patterns in the shared core and the designs that use them.}
\small
\begin{tabular}{>{\raggedright\arraybackslash}p{0.211\linewidth}>{\raggedright\arraybackslash}p{0.458\linewidth}>{\raggedright\arraybackslash}p{0.211\linewidth}}
\toprule
\textbf{Entity} & \textbf{What the pattern matches} & \textbf{Used by} \\ \midrule
EMAIL\_ADDRESS & local part, @, domain and a top-level domain of two or more letters & Regex, Hybrid, Fast hybrid \\
PROFILE\_URL & linkedin.com/in/\ldots{} or github.com/\ldots{}, with optional scheme and www & Regex, Hybrid, Fast hybrid \\
US\_SSN & three, two and four digits joined by hyphens & Regex, Hybrid, Fast hybrid \\
PHONE\_NUMBER & ten-digit US number with optional +1, brackets, dots, dashes or spaces & Regex, Hybrid, Fast hybrid \\
STREET\_ADDRESS & house number, optional direction, one to three capitalized words, street suffix, optional unit & All four \\
US\_ZIP & five- or nine-digit ZIP after a two-letter state code (only the ZIP is replaced) & All four \\
DATE\_OF\_BIRTH & date after ``DOB'' or ``Date of birth'' (only the date is replaced) & All four \\
\bottomrule
\end{tabular}
\end{table}

The second part is the known-entity rule. The redactor reads the name of the known entity, which in our example comes from the KNOWN\_SUBJECT line, and searches the whole text, ignoring case, for the full name and for each part of it that is at least two characters long, starting with the longest terms. In the example this one rule catches the uppercase name at the top of a resume and the first name used later in the summary. It is cheap and exact, and it covers the person who appears most often in the text.

The third part merges spans and writes placeholders. Spans from all detectors are sorted by start position, longer spans first, and any span that overlaps one already taken is skipped. Each distinct value, compared in lower case, gets its own numbered token for its entity type, so ``ANDRE KOWALCZYK'' and ``Andre Kowalczyk'' both become [PERSON\_1], while ``Andre'' on its own becomes [PERSON\_2]. This consistency matters to the LLM, which can still see that the same person appears twice and that the reference is somebody else. Plain masking with a single [REDACTED] token would hide exactly that.

\subsection{Regex Redactor}
The regex redactor uses the known-entity rule and all seven patterns, and nothing else. It loads in under a millisecond and never needs a model file. Its weakness, however, is built into the design: a name that does not belong to the known entity, such as a reference in a resume or a colleague copied on an email, matches no pattern and passes straight through to the model.

\subsection{Presidio Redactor}
The Presidio redactor runs Presidio's analyzer engine on the spaCy \texttt{en\_core\_web\_lg} model and asks for six entity types (PERSON, EMAIL\_ADDRESS, PHONE\_NUMBER, US\_SSN, CREDIT\_CARD and URL) with a score threshold of 0.4. Presidio has no built-in recognizers for US street addresses, ZIP codes or dates of birth in the form we need, so the three shared patterns are added on top, along with the known-entity rule. Two small fixes turned out to be necessary. First, spaCy sometimes lets a person entity run across a line break, for example from the name header into the address below it, so each PERSON span is cut at the first newline. Second, spaCy tags some domain words as people, in our example section headings or ``jack'' in ``pallet jack'', so a short stop-list drops any PERSON span that contains one of them. Each project needs its own stop-list for the same reason.

\subsection{Hybrid Redactor}
The hybrid redactor keeps Presidio but asks it for PERSON only. Emails, phone numbers, URLs, SSNs and addresses come from the regex set, which already handles them well. The idea is simple: use the model only where a model is really needed, and save the cost of Presidio's other recognizers.

\subsection{Fast Hybrid Redactor}
The fast hybrid takes this idea one step further and drops Presidio altogether. It loads the same spaCy model but disables every component except NER, namely the shared token-to-vector layer, the tagger, the parser, the attribute ruler, the lemmatizer and the sentence segmenter. Person entities are read straight from the spaCy document, with the same newline cut and stop-list as before, and everything else comes from regex. Since the parser and the tagger do a lot of work that PERSON detection never uses, removing them should cut most of the cost per document.

\section{Experimental Setup}
\subsection{Synthetic Corpus}
For the ProfileAssist example we used synthetic data, so that the ground truth is exact and no real personal data is involved. A generator with a fixed random seed produced 100 resumes for a warehouse and logistics role, a common entry-level profile. Each resume combines a first and a last name, each drawn from ten options, one of five reference names and one of three US city, state and ZIP combinations. The names are deliberately varied in origin (Kowalczyk, Haddad, Okafor and Lindqvist, for example), because NER models often do worse on names that are rare in their training data. Every document follows the layout shown later in Example 1, with a KNOWN\_SUBJECT line, contact details, a summary that repeats the first name, experience, skills and one reference. The average length is 523 characters, which gives 900 scored PII values (nine per document) and 600 business terms that must survive (six per document). The same script can also redact the input column of an exported trace file, but that mode has no ground truth, so we used it only for spot checks.

\subsection{Metrics}
We measure both accuracy and speed. For accuracy, a PII value counts as leaked if it can still be found in the redacted text as a whole word, ignoring case. With L leaked values out of P scored PII values, recall, which we call ``PII caught'', is

\begin{equation}
\text{PII caught} = 1 - \frac{L}{P}
\end{equation}

Retention measures the opposite failure. With K business terms still present, exactly as written, out of B terms in total, the share kept is

\begin{equation}
\text{Kept} = \frac{K}{B}
\end{equation}

For speed, we time each call to the redactor with Python's high-resolution \texttt{perf\_counter} and summarize the 300 samples as the median (P50) and the 95th percentile (P95). Load time is the time taken to build the redactor, including any model loading. Finally, to judge whether the cost matters, we divide the redactor's median by the median latency of the LLM call:

\begin{equation}
\text{LLM share} = \frac{P50_{\text{redactor}}}{P50_{\text{LLM}}}
\end{equation}

\subsection{Procedure}
All four designs ran one after another in a single Python process on one developer workstation, in the order regex, Presidio, hybrid and fast hybrid, using the CPU only. For each design the script first timed the constructor, then redacted all 100 documents three times while timing each call, and finally ran one more pass to score recall and retention, giving 300 timed calls per design.

\section{Results}
\subsection{Accuracy}
Table 3 shows the main results. The three designs that use a model caught every one of the 900 PII values, while regex caught 800 of them, or 88.9\%. All four designs kept all 600 business terms, so none of them over-redacted on this corpus. That last point is worth noting, because the model-based designs would have masked ``pallet jack'' without the stop-list described in Section 4.3.

\begin{table}[H]
\centering
\caption{Load time, latency and accuracy of the four designs (100 documents, 300 timed calls each).}
\small
\begin{tabular}{>{\raggedright\arraybackslash}p{0.167\linewidth}>{\raggedright\arraybackslash}p{0.114\linewidth}>{\raggedright\arraybackslash}p{0.106\linewidth}>{\raggedright\arraybackslash}p{0.106\linewidth}>{\raggedright\arraybackslash}p{0.132\linewidth}>{\raggedright\arraybackslash}p{0.114\linewidth}>{\raggedright\arraybackslash}p{0.141\linewidth}}
\toprule
\textbf{Design} & \textbf{Load (ms)} & \textbf{P50 (ms)} & \textbf{P95 (ms)} & \textbf{PII caught} & \textbf{Kept} & \textbf{LLM share} \\ \midrule
Regex & 0.8 & 0.20 & 0.49 & 88.9\% & 100.0\% & 0.002\% \\
Presidio & 5,021.2 & 37.59 & 49.96 & 100.0\% & 100.0\% & 0.41\% \\
Hybrid & 1,416.7 & 30.23 & 43.42 & 100.0\% & 100.0\% & 0.33\% \\
Fast hybrid & 1,698.6 & 13.48 & 17.10 & 100.0\% & 100.0\% & 0.15\% \\
\bottomrule
\end{tabular}
\end{table}

The 100 values that regex missed are easy to explain once the result is broken down by PII type. Regex caught all 100 subject first names, last names, emails, phone numbers, addresses, ZIP codes, dates of birth and reference phone numbers, and it missed all 100 reference names. The subject is covered by the known-entity rule and the structured fields are covered by patterns, but a reference such as ``Greta Fontaine'' is neither known in advance nor shaped like anything a pattern can match. The reference's phone number was still caught, which makes the leak a little less harmful, but a name together with a job title and an employer is often enough to find a person.

\subsection{Latency}
Figure 2 plots P50 and P95 latency on a log scale. Regex was about two orders of magnitude faster than any model-based option, at 0.20 ms median and 0.49 ms P95. Presidio was the slowest, at 37.59 ms and 49.96 ms. Limiting Presidio to PERSON cut its median by about 20\%, to 30.23 ms, which shows that Presidio's other recognizers do cost something, although most of the time is still spent inside the full spaCy pipeline. The fast hybrid, which runs only the NER component, reached 13.48 ms median and 17.10 ms P95. That is 2.8 times faster than Presidio and 2.2 times faster than the hybrid, with the same recall. Its tail was also tight, since the gap between P50 and P95 was 3.6 ms against 12.4 ms for Presidio.

\begin{figure}[H]
\centering
\begin{tikzpicture}
\begin{axis}[ybar, width=0.82\linewidth, height=5.6cm, bar width=13pt, ymode=log, log origin=infty,
  ymin=0.1, ymax=300, symbolic x coords={Regex,Presidio,Hybrid,Fast hybrid}, xtick=data,
  ylabel={Latency per document (ms, log scale)}, legend style={at={(0.02,0.98)}, anchor=north west, draw=none},
  nodes near coords, every node near coord/.append style={font=\scriptsize},
  point meta=rawy, enlarge x limits=0.18, axis lines*=left]
\addplot[fill=blue!55, draw=black] coordinates {(Regex,0.20) (Presidio,37.59) (Hybrid,30.23) (Fast hybrid,13.48)};
\addplot[fill=blue!15, draw=black] coordinates {(Regex,0.49) (Presidio,49.96) (Hybrid,43.42) (Fast hybrid,17.10)};
\legend{P50, P95}
\end{axis}
\end{tikzpicture}
\caption{Median (P50) and 95th-percentile (P95) redaction latency per document. Note the log scale.}
\end{figure}
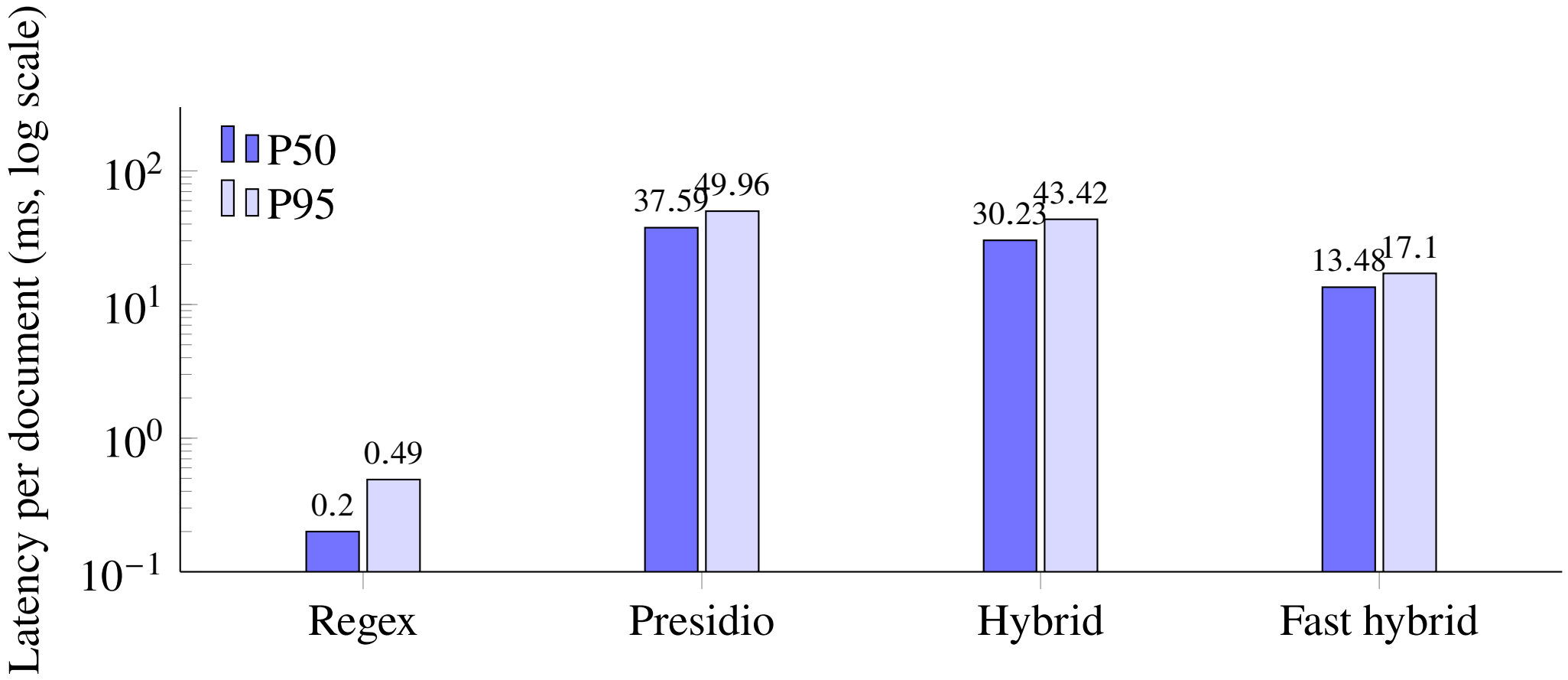

Seen as throughput on a single core, the fast hybrid can process about 74 documents of this size per second, against about 27 for Presidio, a gap that matters more in batch jobs than in a single request.

\subsection{Cost Relative to the LLM Call}
For a design decision, the last column of Table 3 matters most. Even Presidio, the slowest option, adds only 0.41\% to a request whose LLM step takes 9,120 ms at the median. The fast hybrid adds 0.15\%, and regex adds about 0.002\%, which is effectively zero. Reading the accuracy and latency columns together, regex is fast but leaky, while the three model-based designs all reach 100\%, with the fast hybrid the cheapest of them. Put plainly, the step from 88.9\% to 100\% recall costs about 13 ms on a request that already takes more than nine seconds.

\subsection{Load Time}
Load time tells a different story. Regex compiled its patterns in 0.8 ms, whereas Presidio took 5,021 ms to start, the hybrid 1,417 ms and the fast hybrid 1,699 ms. These costs are paid once per process rather than once per request, so in a long-running service they hardly matter. They do matter for serverless functions with frequent cold starts, where a five-second start-up would be felt by users.

\begin{center}\small\textbf{Example 1:} Fast hybrid output for one synthetic resume (hybrid and Presidio outputs are the same apart from the URL label).\end{center}
\begin{Verbatim}[frame=single, fontsize=\footnotesize, xleftmargin=1.2cm, xrightmargin=1.2cm]
KNOWN_SUBJECT: [PERSON_1]
<subject_data> === RESUME ===
[PERSON_1]
[STREET_ADDRESS_1], Memphis, TN [US_ZIP_1]
[EMAIL_ADDRESS_1] | [PHONE_NUMBER_1] | [PROFILE_URL_1]
DOB: [DATE_OF_BIRTH_1]

SUMMARY
Warehouse associate in Memphis. [PERSON_2] prefers first shift.

EXPERIENCE
Forklift Operator - Northwind Logistics (2021 - present)
- Operated pallet jack and reach truck; cycle counting in WMS.

SKILLS
forklift operation, OSHA 10, pallet jack

REFERENCES
[PERSON_3], Shift Supervisor - [PHONE_NUMBER_2]
</subject_data>
\end{Verbatim}

\subsection{A Worked Example}
Example 1 shows the fast hybrid output for one synthetic resume. The subject's full name appears twice and in two different cases, yet both map to [PERSON\_1], and the first name on its own maps to [PERSON\_2]. The reference becomes [PERSON\_3], clearly a different person, while city, state, employer and skills all stay readable. The regex output for the same document is identical except for the last line, which reads ``Greta Fontaine, Shift Supervisor - [PHONE\_NUMBER\_2]''. The Presidio output differs in one further detail: the LinkedIn path is labeled [URL\_1] by Presidio's generic URL recognizer instead of [PROFILE\_URL\_1].

\section{Discussion}
\subsection{Why Regex Alone Is Not Enough}
At first sight regex looks attractive. It is fast, it needs no model and it is easy to audit. Our results, however, show its limit clearly. In most real text the known entity is only one of the people mentioned. A resume names references and former managers, a support ticket names agents and family members, and an email thread names everyone in copy. None of these names are known in advance, and none of them follow a pattern, so a regex-only design will leak them every time. Making the regex cleverer, for instance by matching any two capitalized words, would quickly start to mask job titles, company names and place names, which is precisely the over-redaction that the retention metric is there to catch.

\subsection{Why the Fast Hybrid}
The fast hybrid gives the model only the job a model is good at, which is finding names, and leaves structured identifiers to patterns that are exact and cheap. Because it calls spaCy's NER directly, it avoids both Presidio's extra recognizers and the parts of the spaCy pipeline that PERSON detection does not use. The result is full recall at roughly a third of Presidio's cost, with fewer moving parts: one model and a list of patterns instead of an analyzer framework with its own registry, context logic and dependencies.

Presidio still has its place. It offers checksum validation, context-aware scoring, more languages and a ready-made anonymizer, and teams that need those features may well accept its higher cost, which in our setting was still under half a percent of the LLM call.

\subsection{Practical Recommendations}
Based on these results, we suggest six practices for teams that run LLM pipelines over personal data. First, redact before the LLM call and not only before logging, so that the provider, the logs and the traces all see the same safe text. Second, use a known-entity rule whenever the system already knows whose data it is handling, since it is the cheapest high-recall control available. Third, use NER for person names only and leave structured identifiers to regex. Fourth, load the model once at process start and keep it warm, so that the load cost is never paid per request. Fifth, measure both leakage and retention on a labeled set before every change to patterns or models, and track redaction latency next to LLM latency in the same trace. Sixth, keep regex as a fallback: if the NER model fails to load, a service that falls back to regex-only redaction leaks some names, but a service that skips redaction leaks everything.

\subsection{Applying the Approach to Other Projects}
Nothing in the four designs is specific to resumes. To reuse them in another project, a team changes three things. First, the source of the known entity, such as the customer behind a ticket, the employee behind an HR request or the patient behind a clinical note. Second, the pattern set, which should match the identifier formats of the project's region and domain. Third, the stop-list and the list of business terms that must survive, which are what the retention metric checks. The evaluation method stays the same: build a small labeled set that looks like the project's prompts, measure PII caught and terms kept, and compare redaction latency with the LLM latency from the project's own traces.

\section{Threats to Validity}
Several limits apply to these findings. First, the results come from one sample workflow, and the corpus is synthetic, short and uniform. Real text is longer, often several thousand characters, and varies widely in layout, language and the kinds of PII it holds, so the 100\% recall of the model-based designs should be read as a result on this corpus and not as a general claim; other domains need their own labeled set before the numbers can be carried over. On messy real text we expect some misses from all of the designs. Second, the name pool is small, so the test does not probe how NER copes with rare names, initials or names that are also common words. Third, profile URLs and SSNs are redacted but not scored. Fourth, all timings came from one workstation and one process, with the designs run in a fixed order. Presidio was loaded first, so the later designs benefited from modules already imported and model files already cached, which probably flatters their load times. Finally, the 9,120 ms LLM median comes from one workload over one period. A faster model would raise the relative cost of redaction; even so, if spaCy's cost grows roughly in line with text length, the fast hybrid on a document ten times longer would take around 135 ms, which is still about 1.5\% of the current LLM median.

\section{Conclusion and Future Work}
This paper compared four ways to remove PII from text before it reaches an LLM, using ProfileAssist, a sample resume-to-profile workflow, as a worked example. Regex alone was by far the fastest, but it missed every name that was not known in advance and caught 88.9\% of PII. Presidio, a Presidio-plus-regex hybrid and a fast hybrid built on spaCy's NER component all caught 100\% of PII and kept 100\% of business terms. Among these, the fast hybrid was the quickest at a 13.48 ms median, 2.8 times faster than Presidio, and it added only 0.15\% to a 9,120 ms LLM call. The broader lesson, which holds beyond this example, is that the cost of redaction should be judged against the call it protects; seen that way, a few extra milliseconds are a small price for not sending personal data to a third party. Future work will apply the same benchmark to other LLM workloads that handle personal data, run it on de-identified trace exports with human-labeled ground truth, extend it to longer and non-US text, add transformer-based NER models to the comparison, and test whether placeholder tokens change the quality of the LLM's output.

\clearpage
\bibliographystyle{ieeetr}
\bibliography{references}

@misc{gdpr,
  author       = {{European Parliament and Council of the European Union}},
  title        = {Regulation ({EU}) 2016/679 ({General Data Protection Regulation})},
  howpublished = {Official Journal of the European Union, L 119, pp. 1--88},
  year         = {2016}
}

@misc{ccpa,
  author       = {{State of California}},
  title        = {{California Consumer Privacy Act of 2018}},
  howpublished = {Cal. Civ. Code \S{} 1798.100 et seq.},
  year         = {2018}
}

@misc{dpdp,
  author       = {{Government of India}},
  title        = {The {Digital Personal Data Protection Act}, 2023},
  howpublished = {Act No. 22 of 2023, Ministry of Law and Justice, New Delhi},
  year         = {2023}
}

@inproceedings{carlini,
  author    = {Carlini, N. and others},
  title     = {Extracting training data from large language models},
  booktitle = {Proc. 30th USENIX Security Symposium},
  pages     = {2633--2650},
  year      = {2021}
}

@inproceedings{lukas,
  author    = {Lukas, N. and Salem, A. and Sim, R. and Tople, S. and Wutschitz, L. and Zanella-B{\'e}guelin, S.},
  title     = {Analyzing leakage of personally identifiable information in language models},
  booktitle = {Proc. IEEE Symp. Security and Privacy (SP)},
  pages     = {346--363},
  year      = {2023}
}

@misc{owasp,
  author       = {{OWASP Foundation}},
  title        = {{OWASP Top 10 for Large Language Model Applications}},
  howpublished = {version 2025, OWASP GenAI Security Project},
  year         = {2025}
}

@misc{presidio,
  author       = {{Microsoft}},
  title        = {Presidio: Data protection and de-identification {SDK}},
  howpublished = {GitHub repository. [Online]. Available: \url{https://github.com/microsoft/presidio}}
}

@misc{spacy,
  author       = {Honnibal, M. and Montani, I. and Van Landeghem, S. and Boyd, A.},
  title        = {{spaCy}: Industrial-strength natural language processing in {Python}},
  howpublished = {Zenodo, doi: 10.5281/zenodo.1212303},
  year         = {2020}
}

@inproceedings{sweeney,
  author    = {Sweeney, L.},
  title     = {Replacing personally-identifying information in medical records, the {Scrub} system},
  booktitle = {Proc. AMIA Annual Fall Symposium},
  pages     = {333--337},
  year      = {1996}
}

@article{neamatullah,
  author  = {Neamatullah, I. and others},
  title   = {Automated de-identification of free-text medical records},
  journal = {BMC Medical Informatics and Decision Making},
  volume  = {8},
  note    = {Art. no. 32},
  year    = {2008}
}

@article{stubbs,
  author  = {Stubbs, A. and Kotfila, C. and Uzuner, {\"O}.},
  title   = {Automated systems for the de-identification of longitudinal clinical narratives: Overview of 2014 {i2b2/UTHealth} shared task {Track 1}},
  journal = {Journal of Biomedical Informatics},
  volume  = {58},
  note    = {suppl.},
  pages   = {S11--S19},
  year    = {2015}
}

@article{dernoncourt,
  author  = {Dernoncourt, F. and Lee, J. Y. and Uzuner, {\"O}. and Szolovits, P.},
  title   = {De-identification of patient notes with recurrent neural networks},
  journal = {Journal of the American Medical Informatics Association},
  volume  = {24},
  number  = {3},
  pages   = {596--606},
  year    = {2017}
}

@inproceedings{lample,
  author    = {Lample, G. and Ballesteros, M. and Subramanian, S. and Kawakami, K. and Dyer, C.},
  title     = {Neural architectures for named entity recognition},
  booktitle = {Proc. NAACL-HLT},
  pages     = {260--270},
  year      = {2016}
}

@inproceedings{lison,
  author    = {Lison, P. and Pil{\'a}n, I. and S{\'a}nchez, D. and Batet, M. and {\O}vrelid, L.},
  title     = {Anonymisation models for text data: State of the art, challenges and future directions},
  booktitle = {Proc. 59th Annu. Meeting of the Association for Computational Linguistics (ACL-IJCNLP)},
  pages     = {4188--4203},
  year      = {2021}
}

@article{tab,
  author  = {Pil{\'a}n, I. and Lison, P. and {\O}vrelid, L. and Papadopoulou, A. and S{\'a}nchez, D. and Batet, M.},
  title   = {The text anonymization benchmark ({TAB}): A dedicated corpus and evaluation framework for text anonymization},
  journal = {Computational Linguistics},
  volume  = {48},
  number  = {4},
  pages   = {1053--1101},
  year    = {2022}
}

@inproceedings{yermilov,
  author    = {Yermilov, O. and Raheja, V. and Chernodub, A.},
  title     = {Privacy- and utility-preserving {NLP} with anonymized data: A case study of pseudonymization},
  booktitle = {Proc. 3rd Workshop on Trustworthy Natural Language Processing (TrustNLP)},
  year      = {2023}
}

@techreport{nist122,
  author      = {McCallister, E. and Grance, T. and Scarfone, K.},
  title       = {Guide to protecting the confidentiality of personally identifiable information ({PII})},
  institution = {National Institute of Standards and Technology},
  type        = {NIST Special Publication},
  number      = {800-122},
  address     = {Gaithersburg, MD, USA},
  year        = {2010}
}

@techreport{nistrmf,
  author      = {{National Institute of Standards and Technology}},
  title       = {Artificial Intelligence Risk Management Framework ({AI RMF} 1.0)},
  institution = {National Institute of Standards and Technology},
  type        = {NIST AI},
  number      = {100-1},
  address     = {Gaithersburg, MD, USA},
  month       = jan,
  year        = {2023}
}

@book{friedl,
  author    = {Friedl, J. E. F.},
  title     = {Mastering Regular Expressions},
  edition   = {3rd},
  publisher = {O'Reilly Media},
  address   = {Sebastopol, CA, USA},
  year      = {2006}
}

\end{document}